%% file: main.tex
\documentclass[runningheads]{llncs}

\usepackage{amssymb}
\usepackage{graphicx}
\usepackage{url}
\usepackage{color}
\usepackage{amsfonts}
\usepackage{blindtext}
\usepackage{mathrsfs}
\usepackage{bbm}
\usepackage{url}
\usepackage{algorithm}
\usepackage{algorithmic}
\usepackage{paralist}
\usepackage[english]{babel}
\usepackage{booktabs}
\usepackage{multicol}
\usepackage{multirow}
\usepackage{subfig}
\usepackage{ulem}

\begin{document}
\title{Neural Multi-Task Learning for Teacher Question Detection in Online Classrooms}

\author{Gale Yan Huang\inst{1}\inst{3} \and Jiahao Chen\inst{1} \and Haochen Liu\thanks{Work was done when the author did internship in TAL Education Group}\inst{2} \and Weiping Fu\inst{1} \and Wenbiao Ding\inst{1} \and Jiliang Tang\inst{2} \and  Songfan Yang\inst{1} \and Guoliang Li\inst{3} \and Zitao Liu\thanks{Corresponding Author: Zitao Liu}\inst{1}}

\authorrunning{G. Huang et al.}

\institute{TAL Education Group, Beijing, China  \\
\email{\{galehuang,chenjiahao,fuweiping,dingwenbiao,yangsonogfan,liuzitao\}@100tal.com}
\and Data Science and Engineering Lab, Michigan State University, USA\\
\email{\{liuhaoc1,tangjili\}@msu.edu}
\and Department of Computer Science, Tsinghua University, Beijing, China\\
\email{liguoliang@tsinghua.edu.cn}
}
\maketitle 

\begin{abstract}
\input{abstract}
\end{abstract}

\section{Introduction}
\label{sec:intro}
\input{intro}

\section{Related Work}
\label{sec:related}
\input{related}

\section{Problem Statement}
\label{sec:problem}
\input{problem}

\section{The Proposed Framework}
\label{sec:method}
\input{method}

\section{Experiment}
\label{sec:experiment}
\input{experiment}

\section{Conclusion}
\label{sec:conclusion}
\input{conclusion}

\section*{Acknowledgements}
Haochen Liu and Jiliang Tang are supported by the National Science Foundation of United States under IIS1714741, IIS1715940, IIS1715940, IIS1845081 and IIS1907704. 

%
%
%
%
\bibliographystyle{splncs04.bst}
\bibliography{aied2020}
\end{document}

%% file: abstract.tex
Asking questions is one of the most crucial pedagogical techniques used by teachers in class. It not only offers open-ended discussions between teachers and students to exchange ideas but also provokes deeper student thought and critical analysis. Providing teachers with such pedagogical feedback will remarkably help teachers improve their overall teaching quality over time in classrooms. Therefore, in this work, we build an end-to-end neural framework that automatically detects questions from teachers' audio recordings. Compared with traditional methods, our approach not only avoids cumbersome feature engineering, but also adapts to the task of multi-class question detection in real education scenarios. By incorporating multi-task learning techniques, we are able to strengthen the understanding of semantic relations among different types of questions. We conducted extensive experiments on the question detection tasks in a real-world online classroom dataset and the results demonstrate the superiority of our model in terms of various evaluation metrics.

%% file: intro.tex
Teachers utilize various pedagogical techniques in their classrooms to inspire students' thought and inquiry at deeper levels of students' comprehension. These techniques may include lectures, asking questions, assigning small-group work, etc. \cite{blanchard2016semi,10.1145/3366423.3380018,chen2019multimodal}. A large body of research has demonstrated that asking certain types of questions can increase student engagement and it is an important factor of student achievement \cite{applebee2003discussion,kelly2007classroom,sweigart1991classroom,beck1996questioning,graesser1994question,nystrand1991instructional}. Asking questions has become a central component of teachers' dialogic instructions and often serves as a catalyst for in-depth classroom discussions \cite{kelly2008race,nystrand2003questions,macneilley1998opening}. 

A large spectrum of approaches have been developed and successfully applied in generating classroom feedback to evaluate teachers' performances and help them improve their pedagogical techniques \cite{nystrand2003questions,stivers2010coding,gamoran2003tracking,kane2012gathering,macneilley1998opening}. For example, the Nystrand and Gamoran coding scheme provides a general template for recording teachers' activities, which are used by trained human judges to manually assess teachers' classroom practices \cite{gamoran2003tracking,macneilley1998opening}. However, manually analyzing teacher questions is very subjective, expensive, time-consuming and not scalable. Thus, it is crucial to develop computational methods that can automatically detect teacher questions in live classrooms. By automatically analyzing when teachers ask questions and the corresponding question types, we are able to evaluate the question impact on teaching achievements and help teachers make adjustments to improve their pedagogical techniques. Previous endeavors have been conducted to tackle this problem using traditional machine learning (ML) algorithms \cite{blanchard2015automatic,blanchard2016identifying,blanchard2016semi,DBLP:conf/lak/DonnellyBOKND17,DBLP:conf/edm/SameiOKNDBSGG14}. However, the majority of these methods are not sufficient for teacher question detection due to the following challenges:

\begin{itemize}

\item \textit{Question type variation}. Different from questions in daily chatting, routine conversation or other scenarios, teacher questions in classrooms are very diverse and open-ended. There are different types of classroom questions, such as knowledge-solicitation questions (e.g., ``\textit{What's the definition of quadrangle?}''), open questions (e.g., ``\textit{Could you tell me your thought?}''), procedural questions (e.g., ``\textit{Can everyone hear me?}''), and discourse-management questions (e.g., ``\textit{What?}'', ``\textit{Excuse me?}'') \cite{blosser1991ask,nystrand2003questions}. Traditional methods fail to perform a deep semantic understanding on natural languages, which is necessary for detecting questions of various types.

\item \textit{Subject and speaker variability}. Teaching materials and styles vary dramatically for different subjects and teachers, which leads to significantly distinguished classroom question sentences. Traditional methods show poor adaptability. When new subjects or teachers appear, most existing approaches have to be redesigned and retrained with the newly arrived data. 

\item \textit{Tedious feature engineering}. Traditional ML-based methods detect questions based on complex acoustic and language features. It's time-consuming to construct manually-engineered patterns.


\end{itemize}

In this work, we aim to investigate accurate teacher question detection in online classrooms. In particular, we study two variants of the teacher question detection problem. One is a two-way detection task that aims to distinguish questions from non-questions. The other is a multi-way detection task that aims to classify different types of questions. Please note that the formal definitions of the above two tasks are introduced in Section \ref{sec:problem_def}. We design a neural natural language understanding (NLU) model to automatically extract semantic features from teachers' sentences for both the two-way task and the multi-way task. Our approach shows a powerful generalization capability for detecting questions of various types from different teaching scenarios. With the neural model as a core component, we build an end-to-end framework that directly takes teacher audio tracks as input and outputs the detection results. Experiments conducted on a real-world online education dataset demonstrate the superiority of our proposed framework compared with competitive baseline models.



%% file: related.tex

\subsection{Teacher Question Detection}
\label{sec:question}

Blanchard et al. explore classifying Q\&A discourse segments based on audio inputs \cite{blanchard2015automatic}. A simple amplitude envelope thresholding-based method is developed to detect teachers' utterances. Then the authors extract 11 speech-silence features from detected utterances and train supervised classifiers to differentiate Q\&A segments from other segments. Following this work, Blanchard et al. further introduce an automatic speech recognition (ASR) system to convert audio features into domain-general language features for teacher question detection \cite{blanchard2016identifying,blanchard2016semi}. They extract 37 NLP features from ASR transcriptions and train different classical ML models to distinguish questions from non-questions. Besides linguistic features, Donnelly et al. try both prosodic and linguistic features for supervised question classification and conclude that ML classifiers can achieve better performance with linguistic features \cite{DBLP:conf/lak/DonnellyBOKND17}. 

The line of research presented above focuses on detecting questions from non-questions, which is a binary classification problem. Besides, we are interested in classifying questions into specific categories. Samei et al. build ML models to predict two properties ``uptake'' and ``authenticity'' of questions in live classrooms \cite{DBLP:conf/edm/SameiOKNDBSGG14}. They extract 30 linguistic features related to part-of-speech and pre-defined keywords from each individual question. Samei et al. show that ML models are able to achieve comparable question detection performance as human experts. 

Different from previous works of building question detection ML models based on manually selected linguistic and acoustic features, our approach eliminates the feature engineering efforts and directly learns effective representations from the ASR transcriptions. Furthermore, we introduce multi-task learning techniques into our model to classify different types of questions.

\subsection{Multi-Task Learning}
\label{sec:multi}

Multi-task learning is a promising learning paradigm that aims at taking advantage of information shared in multiple related tasks to help improve the generalization performance of all tasks \cite{caruana1998multitask}. In multi-task learning, a model is trained with multiple objectives towards different tasks simultaneously, where all or some of the tasks are related. Researches have shown that learning multiple tasks jointly can achieve better performance than learning each task separately. Yang et al. propose a novel multi-task representation learning model that learns cross-task sharing structures at each layer of a neural network \cite{yang2016deep}. Hashimoto et al. propose a joint multi-task model for multiple NLP tasks \cite{DBLP:journals/corr/HashimotoXTS16}. The authors point out that training a single network to model the hierarchical linguistic information from morphology, syntax to semantics can improve its generalization ability. Kendall et al. observe that the performance of the multi-task learning framework heavily depends on the weights of the objectives for different tasks \cite{DBLP:journals/corr/KendallGC17}. They develop a novel method to learn the multi-task weightings by taking the homoscedastic uncertainty of each task into consideration.

%% file: problem.tex
The teacher question detection task in live classrooms identifies questions from teachers' speech and classify those questions into correct categories. In this section, we first introduce the method for coding questions and then formulate the problem of teacher question detection.

\subsection{Question Coding}
\label{sec:questcode}

By analyzing thousands of classroom recordings and surveying hundreds of instructors and educators, we categorize teacher questions into the following four categories:

\begin{itemize}
    \item \textbf{Knowledge-solicitation Question (KQ)}: Questions that ask for a knowledge point or a factual answer. Some examples include: ``What's the solution to this problem?'', ``What's the distance between A and B?'', and ``What is the area of this quadrilateral?''.
    \item \textbf{Open Question (OQ)}: Questions to which no deterministic answer is expected. Open questions usually provoke a cognitive process of students such as explaining a problem and talking about knowledge points. Some Examples are: ``How to solve this problem?'', ``Can you share your ideas?'', and ``Why did you do this problem wrong?''.
    \item \textbf{Procedural Question (PQ)}: Questions that teachers use to manage the teaching procedure, such as testing teaching equipment, greeting students, and asking them something unrelated to course content. Examples are: ``Can you hear me?'', ``How are you doing?'', and ``Have I told you about it?''.
    \item \textbf{Discourse-management Question (DQ)}: Questions that teachers use to manage the discourses, such as making transitions or drawing students' attention. Examples include: ``Right?'', ``Isn't it?'', and ``Excuse me?''.
\end{itemize}

We ask crowdsourcing annotators to code each utterance segment as non-question or one of the above four types of questions. The annotators code utterance segments by listening to the corresponding audio tracks. To ensure the coding quality, we first test the annotators on a set of 400 gold-standard examples. The 400 gold-standard examples are randomly sampled from the dataset and annotated by two experienced specialists in education. We only keep the top five annotators who achieve precision scores over 95\% and 85\% on the two-way and multi-way tasks on the gold-standard set to code the entire dataset.

\subsection{Problem Formulation}
\label{sec:problem_def}

We define the \textbf{two-way task} and the \textbf{multi-way task} for the teacher question detection problem as follows. Let $X=(x_1, \dots, x_n)$ be a transcribed utterance where $x_i$ is the $i$-th word and $n$ is the length of the utterance. In the two-way task, each utterance $X$ is assigned a binary label $Y \in \{Q, NQ\}$ where $Q$ indicates that $X$ is a question and $NQ$ indicates it is not a question. In the multi-way task, each utterance $X$ is assigned a label $Y \in \{KQ,OQ,PQ,DQ,NQ\}$ where $KQ$, $OQ$, $PQ$, $DQ$ indicate that $X$ is a knowledge-solicitation question, open question, procedural question or discourse-management question, respectively, and $NQ$ denotes that $X$ is not a question. Both the two-way task and the multi-way task are treated as classification problems where we seek for predicting the label $Y$ of a given utterance $X$.

%% file: method.tex
In this section, we present our framework for teacher question detection in both two-way and multi-way prediction settings. We first introduce the overview of the proposed framework. After that, we discuss the details of our neural natural language understanding module, which is a key component in our question detection framework.

\subsection{The Framework Overview}
\begin{figure}[!tbph]
\includegraphics[width=\linewidth]{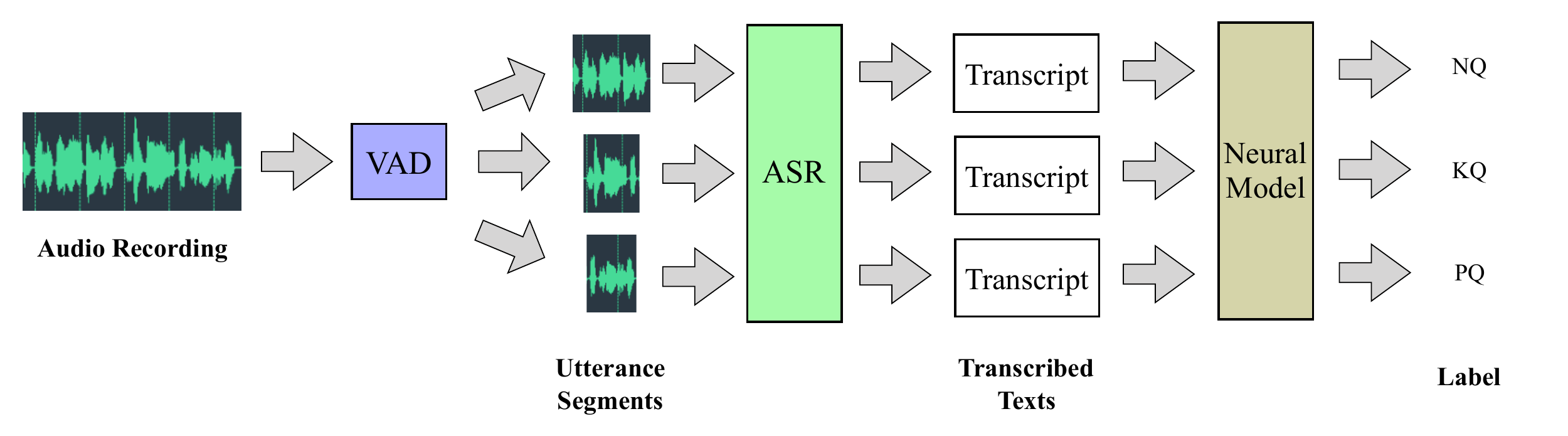}
\caption{The overall workflow of our end-to-end question detection framework.}
\label{fig:framework}
\vspace{-0.5cm}
\end{figure}

The overall workflow of our end-to-end approach is shown in Figure  \ref{fig:framework}. Similar to \cite{li2020multimodal}, we efficiently process the large-volume classroom recordings by utilizing a well-studied voice activity detection (VAD) system to cut an audio recording into small pieces of utterances \cite{sohn1999statistical,zhang2012deep}. The VAD algorithm is able to segment the audio stream into segments of utterances and filter out the noisy and silent ones. Then, each utterance segment is fed into an ASR system for transcription. After that, we build a neural NLU model to extract the semantically meaningful information within each sentence and make the final question detection prediction. Please note that as an end-to-end framework, our model can be integrated seamlessly into a run-time environment in the practical usage.

\subsection{The Neural Natural Language Understanding Model}

In the task of text classification, traditional ML models only use simple word-level features, a.k.a., word embeddings. Due to the fact that such models are not able to extract contextual information, they fail to understand the sentence-level semantics and yield satisfactory detection performance. Therefore, in the work, we propose a neural NLU model to address above issues.

\begin{figure}[!tbph]
\includegraphics[width=\linewidth]{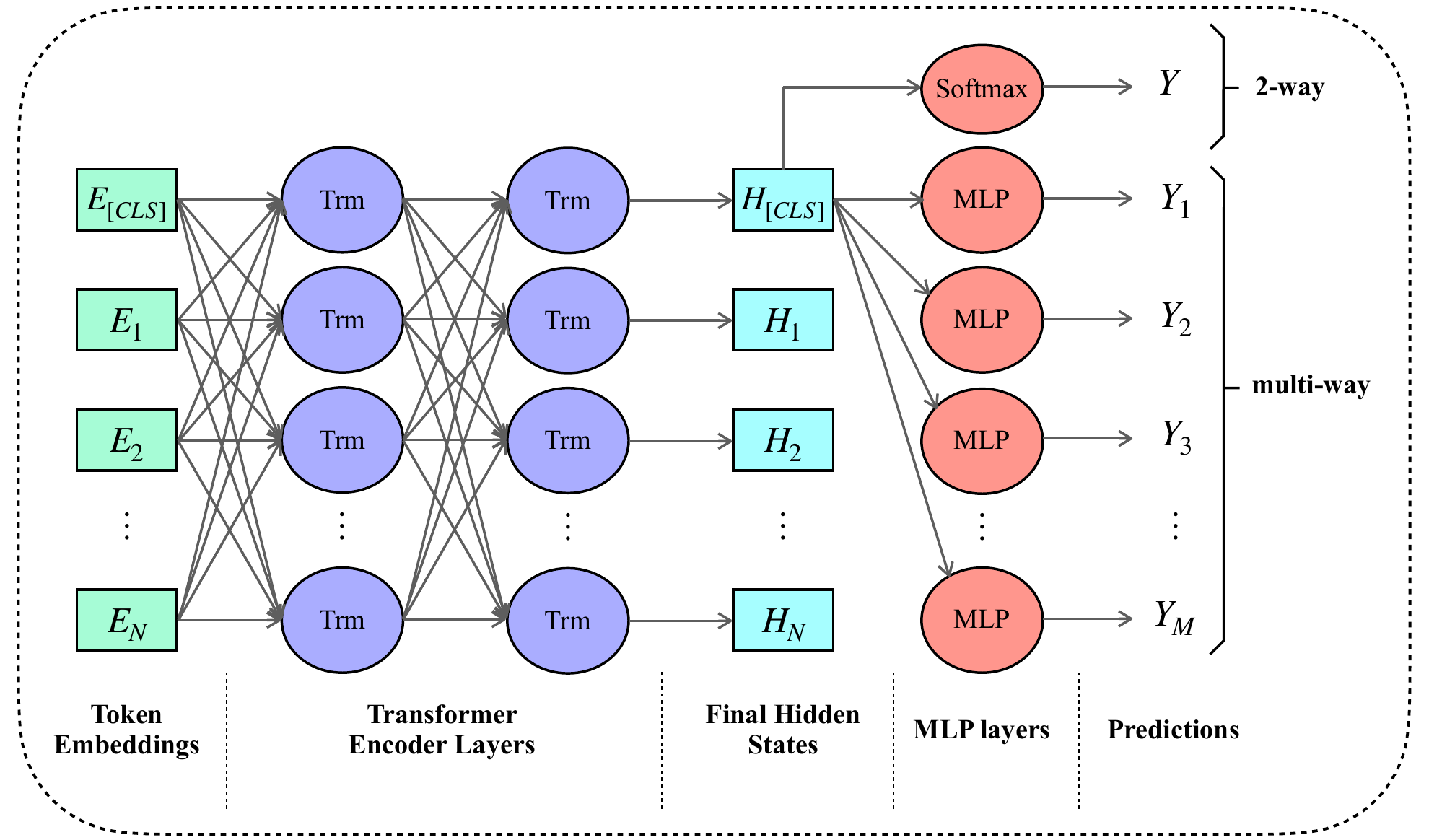}
\caption{An overview of our neural NLU module for question detection.}
\label{fig:model}
\end{figure}

In our NLU module, given a sentence $X=(x_1, \dots, x_N)$ that contains $N$ tokens, similar to Devlin et al. \cite{DBLP:conf/naacl/DevlinCLT19}, we first insert a special token $[CLS]$ in front of the token sequence $X$. Then the sequence of the corresponding token embeddings $E=(E_{[CLS]}, E_1, \dots, E_N)$ is passed through multiple Transformer encoder layers \cite{vaswani2017attention}. Within each Transformer block, each token is repeatedly enriched by the combination of all the words in the sentence so that the contextualized information is captured. At last, we obtain the final hidden states $H=(H_{[CLS]}, H_1, \dots, H_N)$. We treat the final hidden state $H_{[CLS]}$ of the special token $[CLS]$ as the aggregated representation of the entire sentence and use $H_{[CLS]}$ for our two-way and multi-way prediction tasks. Our neural NLU module is shown in Figure \ref{fig:model}. 

\vspace{-0.5cm}

\vspace{-0.5cm}

The NLU structures are different for the two-way and the multi-way tasks. For the two-way task, we feed the final hidden state of the special token $[CLS]$ into a Softmax layer for binary classification. While for the multi-way task, we convert the multi-class classification problem into multiple binary classification problems and train the model in a multi-task learning manner. Suppose that the number of classes is $M$. The final hidden state of $[CLS]$ is fed into $M$ different multi-layer perceptron (MLP) layers to calculate the probabilities of class memberships for each utterance segment. For class $c_i$, the cross-entropy loss function is

$$ L_i = -(\mathbb{I}\{c_i=0\} \log (1-p_i) + \mathbb{I}\{c_i=1\} \log p_i) $$

\noindent where $\mathbb{I}\{\cdot\}$ is an indicator function. $c_i$ is 1 if the utterance segment belongs to the $i$-th class and is 0 otherwise. $p_i$ is the predicted probability that the utterance segment belongs to the $i$-th class. We minimize the sum of cross-entropy loss functions of the $M$ tasks, which is defined as $L_{multi} = \sum_{i=1}^M L_i$. In the inference phase, we make predictions for utterance segments by picking question types with the highest estimated probability.

In this multi-task learning model, multiple binary question classification tasks are learned simultaneously. This method provides several benefits. First, for different tasks, lower layers of the model are shared while the upper layers are different. The shared layers learn to extract the deep semantic features of the input utterance and the upper layers are responsible for making accurate question type predictions. This design yields more modeling capabilities. Second, in teacher question detection, different types of questions share some common patterns, such as interrogative words. But they typically have vastly different contents. When learning a task, the unrelated parts of other tasks can be viewed as auxiliary information, which prevents the model from overfitting and further improves the generalization ability of the model.

%% file: experiment.tex
To verify the effectiveness and superiority of our proposed model, we conduct extensive experiments on a real-world dataset. In this section, we first introduce our dataset that is collected from a real-world online learning platform. Then we describe the details of our experimental setup and the competitive baselines. Finally, we present and discuss the experimental results.

\subsection{Dataset}
\begin{figure}[!tbph]
\includegraphics[width=\linewidth]{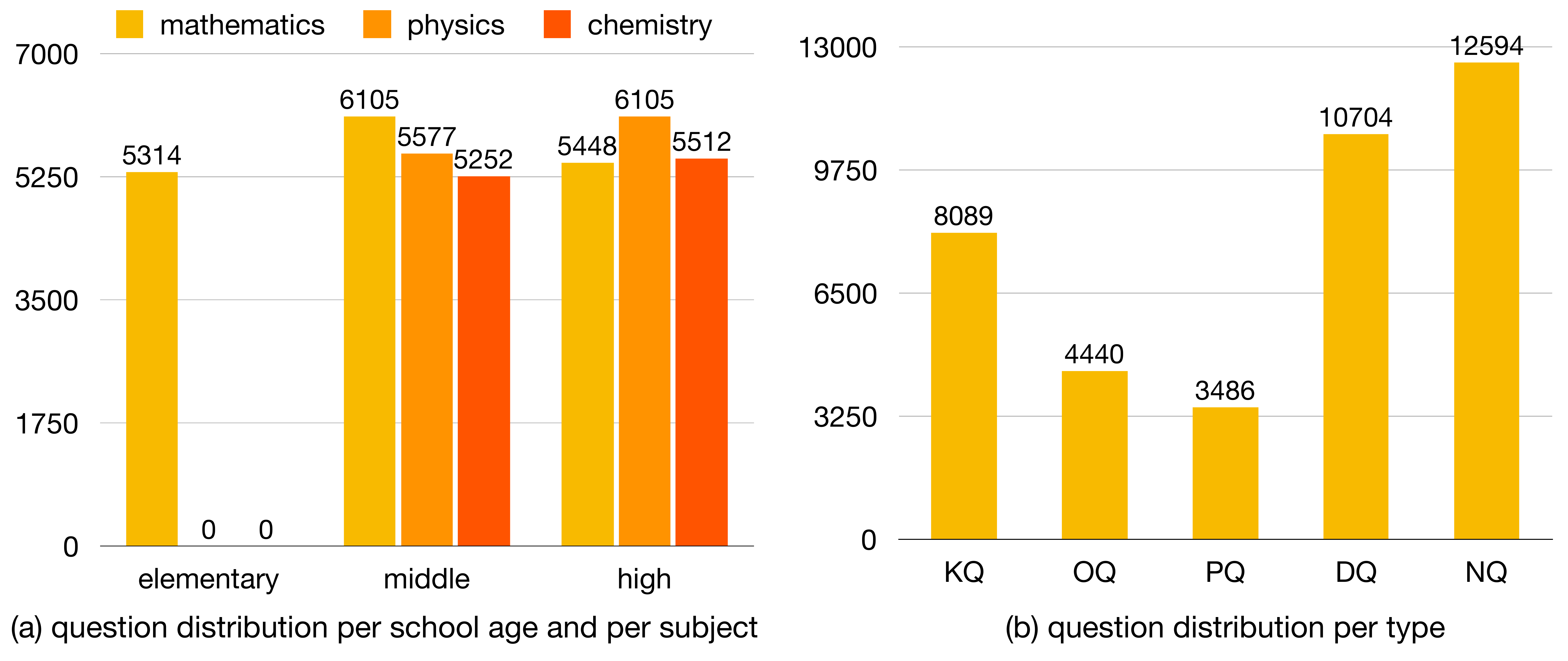}
\caption{Question distributions of our real-world education dataset.}
\label{fig:dist}
\end{figure}

We collect 548 classroom recordings of different subjects and grades from a third party K-12 online education platform\footnote{https://www.xes1v1.com/}. Recordings from both teacher and the students are stored separately in each online classroom. Here, we only focus on the teacher's audio recording. The audio recordings are cut into utterance segments by a self-trained VAD system and each audio segment is transcribed by an ASR service (see Section \ref{sec:detail}). As a result, we obtain 39313 segments in total that are made up of 5314, 16934, and 17065 segments from classes in elementary school, middle school, and high school respectively. The average length of the segments is 3.5 seconds. The detailed segment-level per school-age and per subject question distribution is shown in Figure \ref{fig:dist}(a).

As described in Section \ref{sec:questcode}, each segment is labeled by five qualified annotators. The average pairwise Cohen's Kappa agreement score is 0.696, which indicates a strong annotator agreement. Therefore, we choose to use majority votes as the final labels. The detailed distribution of questions with different types is shown in Figure \ref{fig:dist}(b). We split the whole dataset into a training set, a validation set, and a test set with the proportion of 8:1:1, and the details of data statistics are shown in Table \ref{tab:statistics}.

\begin{table}[!t]
\centering
    \caption{Data statistics of the training set, the validation set, and the test set.}
    \begin{tabular}{l|ccccc|c}
    \hline
    \hline
        \textbf{} & \textbf{KQ} & \textbf{OQ} & \textbf{PQ} & \textbf{DQ} & \textbf{NQ} & \textbf{Total}\\
    \hline
    \textbf{Training} & 6450 & 3551 & 2786 & 8514 & 10149 & 31450\\
    \textbf{Validation} & 861 & 431 & 328 & 1104 & 1207 & 3931\\
    \textbf{Test} & 778 & 458 & 372 & 1086 & 1238 & 3932\\
    \hline
    \hline
    \end{tabular}
    \label{tab:statistics}
    \vspace{-0.3cm}
\end{table}

\subsection{Implementation Details}
\label{sec:detail}

In the work, we train our VAD model by using a four-layer DNN neural network to distinguish normal human utterances from background noises and silences \cite{tashev2016dnn}. Similar to Blanchard et al. \cite{blanchard2015study}, we find that publicly available ASR service may yield inferior performance in the noisy and dynamic classroom environments. Therefore, we train our own ASR models on the classroom specific datasets based on a deep feed-forward sequential memory network proposed by Zhang et al. \cite{zhang2018deep}. Our ASR has a word error rate of 28.08\% in our classroom settings.

Language model pre-training techniques have achieved great improvements on various NLU tasks \cite{DBLP:conf/naacl/DevlinCLT19,DBLP:journals/corr/abs-1909-00204}. In the implementation of our neural NLU model, we first pre-train the model with a large-scale language corpus and then use question specific classroom data to conduct the model fine-tuning. Here, we adopt the pre-trained NEZHA-base model released by Wei et al. \cite{DBLP:journals/corr/abs-1909-00204}. In the multi-task setting, we apply a two-layer MLP with hidden sizes 256, 64 for each class. The output is passed through a sigmoid function to calculate the predictive probability. An optimal set of hyper-parameters is picked according to the model's performance on the validation set and we report its performance on the test set.

\subsection{Baselines}

We compare our approach with the following representative baseline methods: (1) Logistic Regression (LR) \cite{kleinbaum2002logistic}, (2) K-Nearest Neighbor (KNN) \cite{fix1951discriminatory}, (3) Random Forest (RF) \cite{ho1995random}, (4) Support Vector Machine (SVM) \cite{cortes1995support}, (5) Gradient Boosted Decision Tree (GBDT) \cite{drucker1996boosting} and (6) Bidirectional Long Short Term Memory Network (Bi-LSTM) \cite{zhang2015bidirectional}. For the first five baselines, we use the sentence embedding of a given transcribed utterance as the feature vector for classification. The sentence embedding is computed by taking the average of the pre-trained word embeddings within each sentence. For Bi-LSTM, the word embeddings are fed into an LSTM network sequentially and the concatenation vector of the final hidden states in two directions is fed into a Softmax layer for classification. 


\subsection{Experimental Results}

We show the results of the two-way and the multi-way tasks in Table \ref{tab:two-way-task} and Table \ref{tab:multi-way-task}, respectively. In the two-way task, we report the classification results of different models in terms of accuracy, precision, recall, F1 score, and AUC score, respectively. In the multi-way task, we report the classification results on each question type in terms of F1 score, as well as the overall results in terms of precision, recall and F1 scores from both micro and macro perspectives \cite{van2013macro}. From Table \ref{tab:two-way-task} and Table \ref{tab:multi-way-task}, we make the following observations:

\vspace{-0.5cm}

\begin{table}[!bhpt]
\centering
    \caption{Performance comparison of the \textbf{two-way task}.}
    \begin{tabular}{l|c|c|c|c|c}
    \hline
    \hline
        \textbf{} & \textbf{Accuracy} & \textbf{Precision} & \textbf{Recall} & \textbf{F1 Score} & \textbf{AUC}\\
    \hline
    \textbf{LR} & 0.724 & 0.863 & 0.711 & 0.779 & 0.811\\
    \textbf{KNN} & 0.740 & 0.745 & 0.943 & 0.832 & 0.763\\
    \textbf{RF} & 0.766 & 0.758 & \textbf{0.968} & 0.850 & 0.824\\
    \textbf{SVM} & 0.798 & 0.874 & 0.824 & 0.848 & 0.854\\
    \textbf{GBDT} & 0.817 & 0.826 & 0.929 & 0.874 & 0.837\\
    \textbf{Bi-LSTM} & 0.873 & 0.882 & 0.940 & 0.910 & 0.915\\
    \textbf{Our Model} & \textbf{0.885} & \textbf{0.888} & 0.952 & \textbf{0.919} & \textbf{0.933}\\
    \hline
    \hline
    \end{tabular}
    \label{tab:two-way-task}
    \vspace{-0.5cm}
\end{table}

\begin{table}[!bhpt]
\centering
  \caption{Performance comparison of the \textbf{multi-way task}. ma-Pre., ma-Rec., mi-F1 and ma-F1 represent the macro precision, macro recall, micro F1 score and macro F1 score respectively.}
  \begin{tabular}{l|ccccc|cccc}
  \hline
  \hline
    \textbf{Type} & \textbf{KQ} & \textbf{OQ} & \textbf{PQ} & \textbf{DQ} & \textbf{NQ} & \multicolumn{4}{c}{\textbf{Overall}}\\
    \hline
     & \textbf{F1} & \textbf{F1} & \textbf{F1} & \textbf{F1} & \textbf{F1} & \textbf{ma-Pre.} & \textbf{ma-Rec.} & \textbf{mi-F1} & \textbf{ma-F1}\\
    \hline
    \textbf{LR} & 0.621 & 0.584 & 0.532 & 0.734 & 0.620 & 0.611 & 0.634 & 0.637 & 0.618 \\
    \textbf{KNN} & 0.450 & 0.461 & 0.450 & 0.616 & 0.540 & 0.580 & 0.490 & 0.540 & 0.503 \\
    \textbf{RF} & 0.564 & 0.454 & 0.483 & 0.699 & 0.632 & 0.661 & 0.537 & 0.612 & 0.566\\
    \textbf{SVM} & 0.644 & 0.629 & 0.561 & 0.791 & 0.694 & 0.655 & 0.681 & 0.688 & 0.664\\
    \textbf{GBDT} & 0.629 & 0.583 & 0.516 & 0.758 & 0.676 & 0.662 & 0.616 & 0.668 & 0.632\\
    \textbf{Bi-LSTM} & 0.743 & 0.751 & 0.654 & 0.914 & 0.778 & 0.769 & 0.769 & 0.794 & 0.768\\
    \textbf{Our Model} & 0.767 & 0.768 & 0.686 & 0.912 & 0.793 & \textbf{0.781} & \textbf{0.794} & \textbf{0.808} & \textbf{0.785}\\
  \hline
  \hline
  \end{tabular}
  \label{tab:multi-way-task}
  \vspace{-0.3cm}
\end{table}

\begin{itemize}
    \item First, in terms of both the two-way and multi-way tasks and most of the evaluation metrics, our model outperforms all the baseline methods. Due to the fact that our dataset consists of different subjects, school-ages, teachers and question types, we believe that the performance improvements achieved by our approach show the adaptability and robustness towards the real challenging educational scenarios.

    \item Second, by comparing the performances of the models on different types of questions, we find that procedural questions are relatively harder to identify compared to discourse-management questions. We believe the reason is that procedural questions typically involve a wide range of topics and appear in diverse forms. While discourse-management questions are short and succinct, and their forms are relatively fixed.

    \item Third, the baselines LR, KNN, RF, SVM, and GBDT achieve unsatisfactory performance in both tasks. Because they simply average the word embeddings as the features for classification, which fail to capture any contextualized information. Bi-LSTM performs better by learning better contextualized representations. The proposed framework outperforms Bi-LSTM because of its powerful ability of deep semantic understanding learned through the Transformer layers, the pre-training procedure, and the multi-task learning technique.
\end{itemize}

%% file: conclusion.tex
In this paper, we present a novel framework for the automatic detection of teacher questions in online classrooms. We propose a neural NLU model, which is able to automatically extract semantic features from teachers' utterances and adaptively generalize across recordings of different subjects and speakers. Experiments conducted on a real-world education dataset validate the effectiveness of our model in both two-way and multi-way tasks. As a future research direction, we are going to explore the relationship between the use of teacher questions and student achievement in live classrooms, thus we can make corresponding suggestions to teachers to improve their teaching efficiency.